\pdfoutput=1

\documentclass[letterpaper,10pt,journal,twoside]{IEEEtran}
\usepackage{graphicx}
\usepackage{amsmath}
\usepackage{amssymb}
\usepackage{booktabs}
\usepackage{algorithmic}
\usepackage{multirow}
\usepackage[table]{xcolor}
\usepackage{hyperref}
\makeatletter
\newcommand{\captionof}[1]{\def\@captype{#1}\caption}
\makeatother
\usepackage{balance}
\usepackage{needspace}
\hypersetup{colorlinks=true, linkcolor=teal, citecolor=teal, urlcolor=teal,
pdftitle={Swim2Real: Vision-Language Models for Simulator Calibration of a Soft Robotic Fish},
pdfauthor={Kevin Qiu, Kyle Liam Walker, Mike Yan Michelis, Marek Cygan, Josie Hughes}}
\newcommand{\method}{Swim2Real}
\newcommand{\onelinelonger}{\looseness=1\relax}
\newcommand{\bestMeanSd}[1]{\expandafter\bestMeanSdParts#1\relax}
\def\bestMeanSdParts#1\pm#2\relax{\mathbf{#1}\pm#2}
\ifdefined\FINALSUBMISSION
\newcommand{\resultslot}[1]{\PackageError{ral2026}{Unfilled result slot: #1}{Replace every result slot before the final build.}}
\else
\newcommand{\resultslot}[1]{}
\fi

\title{
Swim2Real: Vision-Language Models for Simulator Calibration of a Soft Robotic Fish
}

\author{Kevin Qiu$^{1}$, Kyle Liam Walker$^{2,3}$, Mike Yan Michelis$^{4}$,
Marek Cygan$^{1}$, and Josie Hughes$^{2}$%
\thanks{Manuscript received: June 21, 2026; Revised September 11, 2026; Accepted October 13, 2026.}%
\thanks{This paper was recommended for publication by Editor Barbara Mazzolai upon evaluation of the Associate Editor and Reviewers' comments.
This work was supported in part by the National Science Centre, Poland, under PRELUDIUM Grant 2024/53/N/ST6/03370 and Sonata Bis Grant 2024/54/E/ST6/00388.}%
\thanks{$^{1}$Kevin Qiu and Marek Cygan are with the Faculty of Mathematics, Informatics and Mechanics, University of Warsaw, 02-097 Warsaw, Poland. Correspondence to: {\tt\footnotesize kevinxqiu@gmail.com}}%
\thanks{$^{2}$Kyle Liam Walker and Josie Hughes are with EPFL, 1015 Lausanne, Switzerland.}%
\thanks{$^{3}$Kyle Liam Walker is also with the University of Liverpool, Liverpool L69~7ZX, U.K.}%
\thanks{$^{4}$Mike Yan Michelis is with ETH Zurich, 8092 Zurich, Switzerland.}%
\thanks{Digital Object Identifier (DOI): \href{https://doi.org/10.1109/LRA.2026.3752343}{10.1109/LRA.2026.3752343}.}}

\usepackage{eso-pic}
\begin{document}
\AddToShipoutPictureFG*{\AtPageLowerLeft{\put(\LenToUnit{\dimexpr(\paperwidth-\textwidth)/2\relax},\LenToUnit{0.22in}){\parbox[b]{\textwidth}{\scriptsize\noindent \textcopyright\,2026 IEEE. Personal use of this material is permitted. Permission from IEEE must be obtained for all other uses, in any current or future media, including reprinting/republishing this material for advertising or promotional purposes, creating new collective works, for resale or redistribution to servers or lists, or reuse of any copyrighted component of this work in other works. Accepted for publication in IEEE Robotics and Automation Letters, DOI: \href{https://doi.org/10.1109/LRA.2026.3752343}{10.1109/LRA.2026.3752343}.}}}}
\bstctlcite{IEEEexample:BSTcontrol}

\IEEEaftertitletext{\vspace{-0.5\baselineskip}\centering
\includegraphics[width=\textwidth]{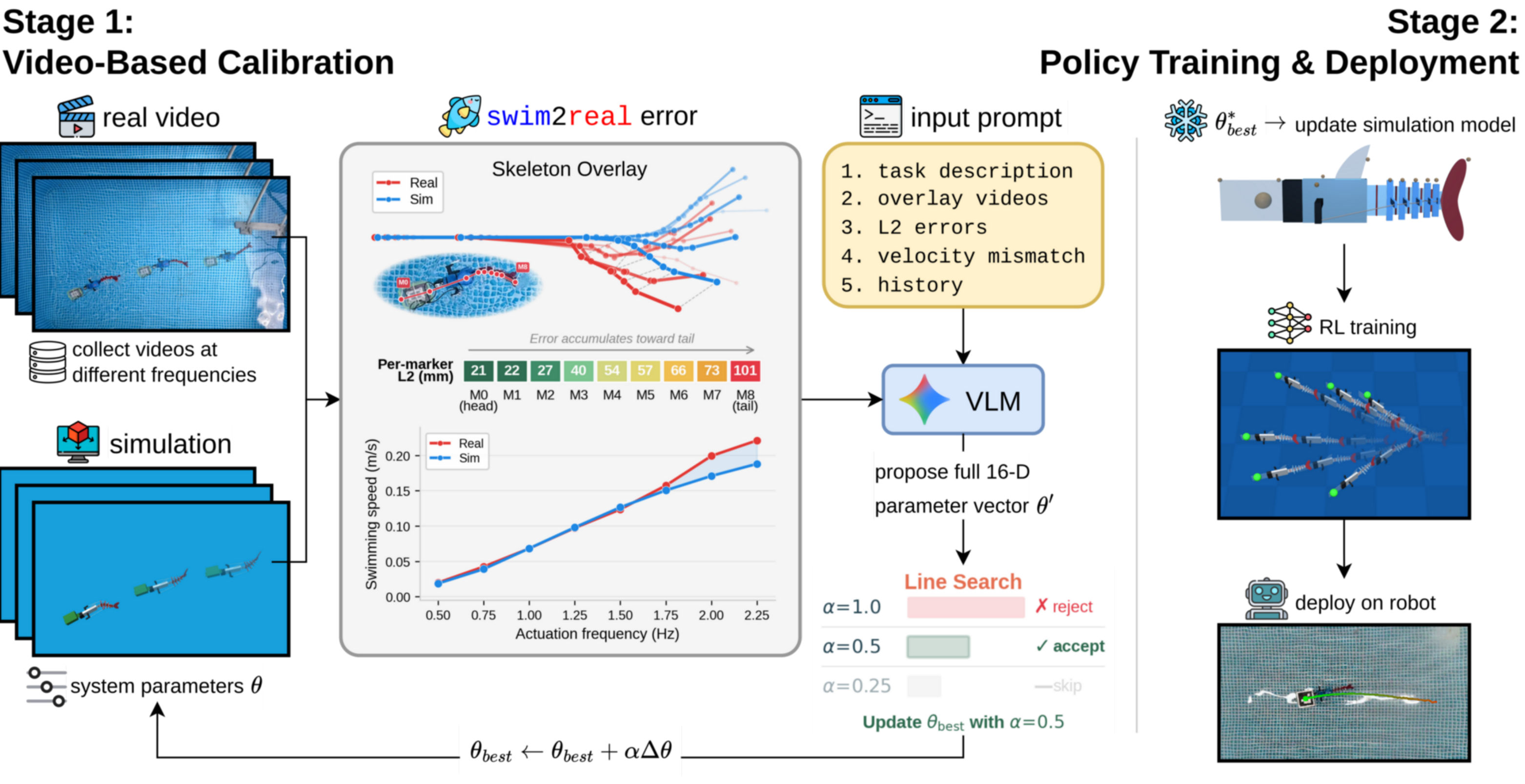}
\captionof{figure}{%
    \method{} workflow. In Stage~1, a vision-language model (VLM) compares simulated and real swimming videos and proposes parameter updates, which a backtracking line search checks against marker error.
    Stage~2 shows reinforcement learning (RL) policy training in simulation followed by open-loop replay of motor commands on the robot. The hardware trials use policies from an earlier simulator and demonstrate command execution separately from the calibration results reported here.
    The insets illustrate the VLM input format.}
\label{fig:framework}
\vspace{0.6\baselineskip}}
\maketitle

\begin{abstract}
Calibrating a soft aquatic robot simulator requires matching coupled deformation and propulsion from limited pool recordings.
\method{} uses a vision-language model (VLM) to compare real and simulated swimming videos and propose updates to 16 simulator parameters. A backtracking line search tests each direction and accepts updates that reduce marker error.
To test whether the proposals carry useful directions, we compare them with random and signed-permuted directions under the same line search and simulator budget.
In an independent, preregistered fifty-seed replication, \method{} reduces the mean best-so-far error over 40 simulator evaluations by 63.4\,mm against random directions and 53.2\,mm against permuted directions. Both paired comparisons pass Holm correction ($p<\mbox{\boldmath$10^{-5}$}$), conditional on run completion.
A 24-seed experiment with a second VLM supports the same direction effect.
Backtracking averages 16.1 VLM proposals per run versus 39 for Vid2Sid, which attains lower error at the same simulator budget.
Ten hardware trials demonstrate open-loop execution of simulator-generated motor commands.
Project page and videos: \mbox{\href{https://swim2real.github.io}{\nolinkurl{swim2real.github.io}}}.
\end{abstract}


\Needspace{5\baselineskip}
\section{Introduction}
\label{sec:introduction}

\IEEEPARstart{R}{obot} learning benefits from simulators that reproduce the dynamics observed on the physical system~\cite{muratore2022robot}.
For soft aquatic robots, calibration must account for distributed elasticity and multibody or continuum dynamics~\cite{della2023model}, together with deformation and thrust generation in water~\cite{zhang2022sim2real}.
A simulator can match one part of a swimming motion while missing another: the tail may bend with the wrong amplitude, or the body may advance too slowly.
A scalar trajectory error measures the mismatch, but does not explain which parameters to change.

\method{} uses the videos themselves as feedback to calibrate a 16-dimensional MuJoCo model of a robotic fish.
A vision-language model (VLM) compares simulated and real swimming videos alongside numerical residuals, then proposes updates to fluid, motor, and joint parameters.
We treat its physical diagnoses as hypotheses and check each proposed direction with a backtracking line search.
The search tries progressively smaller steps until marker error decreases, allowing one VLM proposal to support several simulator evaluations.

We test whether these directions carry useful information beyond what the same line search can extract from random or signed-permuted directions.
A first study uses fifteen shared initializations to compare \method{} and these controls with Bayesian optimization (BO), simultaneous perturbation stochastic approximation (SPSA), and covariance matrix adaptation evolution strategy (CMA-ES) at two population sizes. These optimizers use only the numerical residual. We also compare with Vid2Sid~\cite{qiu2026vid2sid}, the closest video-based VLM method, which requests a fresh proposal for each candidate.
An independent fifty-seed replication supports the direction effect against both controls under the completion rules in Section~\ref{sec:experimental_setup}-D.

The paper contributes:
\begin{enumerate}
    \item Controlled evidence that VLM-proposed directions improve calibration. Under a shared line search and simulator budget, \method{} reduces error relative to random and signed-permuted directions in a fifty-seed replication and with a second VLM.
    \item A video-guided method that turns VLM physical diagnoses into 16-dimensional parameter directions and validates each step with a backtracking line search.
    \item Descriptive evidence at actuation frequencies withheld from fitting. \method{} reaches 53.1\,mm mean assessment error and wins 7/8 paired initializations against random directions and 6/8 against BO.
\end{enumerate}

\section{Related Work}
\label{sec:related_work}

\textbf{Simulator Calibration for Sim-to-Real.}
Domain randomization trains perception or control across varied simulator parameters to tolerate mismatch~\cite{tobin2017domain,peng2018sim,hwangbo2019anymal}, while adaptive variants update the randomization distribution using real rollouts~\cite{chebotar2019closing}.
System identification offers a complementary route by estimating model parameters from observed system behavior~\cite{ljung1999system,ramos2019bayessim}.

\textbf{System Identification for Robotics.}
BayesSim estimates simulator-parameter posteriors through likelihood-free inference~\cite{ramos2019bayessim}.
Generic black-box optimizers such as Bayesian optimization~\cite{snoek2012practical} and CMA-ES~\cite{hansen2016cmaes} can optimize a chosen scalar simulator discrepancy.
Auto-Tuned Sim-to-Real instead learns from observation sequences whether simulator parameters should increase or decrease~\cite{du2021auto}.
Differentiable simulation~\cite{jatavallabhula2021gradsim} provides gradients through simulation, including for soft robotic fish~\cite{zhang2022sim2real}. Du et al.~\cite{du2021underwater} jointly fit material, actuation, and hydrodynamic parameters from recorded underwater robot motion, then optimize open-loop controls.
Soft robotic fish combine compliant structures with underwater propulsion~\cite{marchese2014autonomous,katzschmann2018exploration}, while prior differentiable calibration identifies material parameters from quasi-static measurements and predicts dynamic behavior~\cite{zhang2022sim2real}.

LLMPhy~\cite{cherian2026llmphy} iteratively combines language-model proposals, physics simulation, and reconstruction feedback for physical parameter identification.
Phys2Real~\cite{wang2026phys2real} uses VLM parameter priors together with interaction-based online adaptation for manipulation.
Closest to our video-calibration setting, Vid2Sid~\cite{qiu2026vid2sid} proposes parameter values directly and retains the best observed iterate.

Generic optimizers infer parameter updates from scalar residuals. Differentiable calibration uses simulator gradients, and learned update models require task-specific training data. \method{} uses a pretrained VLM to propose directions without simulator gradients or task-specific training. Reusing each direction across step sizes reduces the number of model queries needed to evaluate candidates.

\textbf{Foundation Models in Robotics.}
Large language models (LLMs) and VLMs have been applied to task planning~\cite{ahn2022saycan}, visuomotor control~\cite{brohan2023rt2}, reward design~\cite{ma2024eureka}, and domain-randomization design~\cite{ma2024dreureka}.
Feedback-driven uses include VLM-based failure diagnosis~\cite{duan2025aha}, task replanning after execution errors~\cite{mei2024replanvlm}, and simulation-in-the-loop action planning~\cite{liu2026simpact}.
Physical-reasoning methods ground VLMs in visual object properties for manipulation~\cite{gao2024physically} and explicit robot-reachability representations~\cite{zhou2025physvlm}.
In optimization, LLMs propose black-box search steps~\cite{yang2024large}, enhance Bayesian optimization surrogates~\cite{liu2024large}, and generate candidates for robot design~\cite{kawaharazuka2025efficient,gao2025vlmgineer}.
These optimization and design-search methods check model-generated proposals through external evaluation, and \method{} applies the same check to low-level simulator calibration, where the marker-error objective validates each video-derived proposal (Fig.~\ref{fig:framework}).

\section{The \method{} Pipeline}
\label{sec:method}

\subsection{Problem Formulation}

Consider a parameterized simulator $\mathcal{S}_\theta$ with parameters $\theta \in \mathbb{R}^d$ and a real robotic system producing reference trajectories.
The goal is to find $\theta^*$ that minimizes discrepancies between simulated and real marker trajectories:
\begin{equation}
    \begin{aligned}[b]
    \theta^* \!= \arg\min_{\theta \in \Theta}\;
    &\frac{1}{|\mathcal{F}|} \sum_{f \in \mathcal{F}} \frac{1}{M T_f}
    \sum_{m=1}^{M} \sum_{t=1}^{T_f} \\
    &\| \tilde{\mathbf{p}}_m^{\mathrm{s}}(t,f,\theta) - \tilde{\mathbf{p}}_m^{\mathrm{r}}(t,f) \|_2
    \end{aligned}
    \label{eq:objective}
\end{equation}
where $\Theta \subset \mathbb{R}^d$ is the bounded parameter space, $\mathcal{F}$ is a set of actuation frequencies, $M$ is the number of body markers, $T_f$ is the number of time steps at frequency $f$, and $\tilde{\mathbf{p}}_m^{\mathrm{s}}, \tilde{\mathbf{p}}_m^{\mathrm{r}} \in \mathbb{R}^2$ are simulated and real marker positions after body-frame and temporal alignment. We denote this objective $\mathcal{L}(\theta)$.
Each evaluation of $\mathcal{L}(\theta)$ requires running $|\mathcal{F}|$ MuJoCo simulations and constitutes the dominant simulator cost.

\subsection{VLM-Guided Direction Proposals}

\begin{figure}[t]
\centering
\includegraphics[width=0.85\columnwidth]{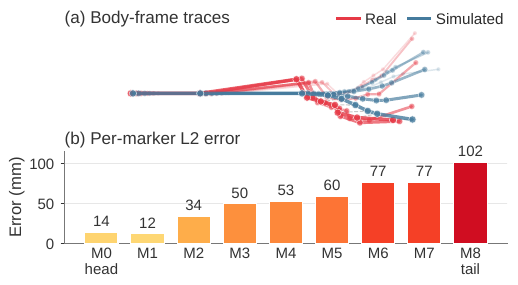}
\caption{Illustrative VLM inputs. Body-frame marker \mbox{trajectories} and per-marker errors at one frequency show the diagnostic format.}
\label{fig:vlm_input_example}
\end{figure}

At each iteration, Gemini 2.5 Pro~\cite{gemini2025gemini25} receives videos and numerical diagnostics of the current sim-real mismatch.
The visual inputs comprise simulated and real marker overlays at each fitting frequency (Fig.~\ref{fig:vlm_input_example}) and a real clip at the worst-matching frequency. Numerical inputs describe per-marker and per-frequency errors and velocities, current parameters and bounds, and prior proposals and outcomes.

The model returns a diagnosis, parameter updates, and confidence. Invalid or out-of-bound proposals are discarded or projected, and omitted parameters remain unchanged. The proposed update sets the direction and full step, while line search chooses the accepted step size.

\subsection{Backtracking Line Search}

\begin{figure}[t]
\centering
\includegraphics[width=0.85\columnwidth]{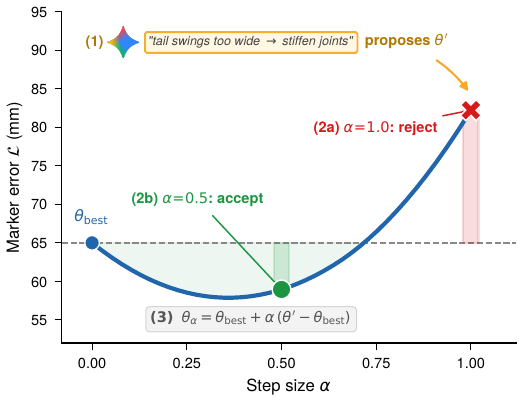}
\caption{Backtracking along a VLM-proposed direction. The full step ($\alpha_0\!=\!1.0$) increases marker error, whereas a smaller step improves on $\mathcal{L}_\mathrm{best}$ and is accepted. The curve is a schematic one-dimensional path through the 16D parameter space.}
\label{fig:linesearch}
\end{figure}

An observed mismatch, such as excessive tail bending, suggests a parameter change but leaves its magnitude to be determined.
We adapt the classical backtracking line-search idea~\cite{nocedal2006numerical} by validating each VLM proposal at decreasing step sizes and accepting the first objective-improving candidate.
Given the current best parameters $\theta_{\text{best}}$ and the VLM proposal $\theta'$, the search direction is $\Delta = \theta' - \theta_{\text{best}}$, and the method evaluates:
\begin{equation}
    \alpha_k = \beta^k,\quad
    \theta_k = \Pi_\Theta(\theta_{\text{best}} + \alpha_k \Delta), \quad k = 0, 1, \ldots, K{-}1
    \label{eq:linesearch}
\end{equation}
where $\beta \in (0,1)$ is the decay factor, $K$ is the maximum number of step sizes to try, and $\Pi_\Theta$ projects candidate parameters to the allowed bounds. The first $k$ yielding $\mathcal{L}(\theta_k) < \mathcal{L}(\theta_{\text{best}})$ is accepted.
If no step improves the objective, the round is rejected and the VLM receives this feedback in the next iteration.
In the 16D experiments we use $\beta = 0.5$ and $K = 3$. The full, half, and quarter steps bound each VLM round to three of the 40 evaluations (Figs.~\ref{fig:linesearch} and~\ref{alg:vlm_golden}).
Every full and backtracked step consumes one of the $B$ simulation evaluations, just as a BO or Vid2Sid candidate does, and VLM proposals are counted separately.
We compare direction reuse descriptively with Vid2Sid~\cite{qiu2026vid2sid}, which requests a fresh proposal for each candidate. Our adaptation uses the shared fish prompt and evaluator, conditioning each proposal on the best-so-far parameters rather than the latest iterate.

\begin{figure}[t]
\footnotesize
\begin{algorithmic}[1]
\REQUIRE Simulator $\mathcal{S}_\theta$, evaluation budget $B$, parameter bounds $\Theta$, decay $\beta$, maximum line-search steps $K$
\STATE Choose initial $\theta_{\text{best}} \in \Theta$ and evaluate $\mathcal{L}_{\text{best}}$;\; $B \leftarrow B{-}1$
\STATE $H \leftarrow \emptyset$ \COMMENT{Optimization history}
\WHILE{$B > 0$}
    \STATE Render sim-real overlay $V$ at $\theta_{\text{best}}$
    \STATE Ask the VLM for proposal $\theta'$ using $V$, $\theta_{\text{best}}$, and $H$
    \STATE $\Delta \leftarrow \theta' - \theta_{\text{best}}$
    \STATE $C \leftarrow \emptyset$;\; $\alpha_\star \leftarrow 0$ \COMMENT{Backtracking line search}
    \FOR{$k = 0, 1, \ldots, K{-}1$}
        \IF{$B = 0$}
            \STATE \textbf{break}
        \ENDIF
        \STATE $\alpha \leftarrow \beta^k$
        \STATE Project $\theta_{\text{best}} {+} \alpha\Delta$ to the bounds $\Theta$ to obtain $\theta_k$
        \STATE Evaluate $\mathcal{L}_k = \mathcal{L}(\theta_k)$;\; $B \leftarrow B{-}1$
        \STATE Record $(\alpha,\mathcal{L}_k)$ in $C$
        \IF{$\mathcal{L}_k < \mathcal{L}_{\text{best}}$}
            \STATE $\theta_{\text{best}} \leftarrow \theta_k$;\; $\mathcal{L}_{\text{best}} \leftarrow \mathcal{L}_k$;\; $\alpha_\star \leftarrow \alpha$
            \STATE \textbf{break}
        \ENDIF
    \ENDFOR
    \STATE Record $(\Delta, C, \alpha_\star, \mathcal{L}_{\text{best}})$ in $H$
\ENDWHILE
\RETURN $\theta_{\text{best}},\; \mathcal{L}_{\text{best}}$
\end{algorithmic}
\caption{\method{}: VLM-guided calibration with line search.}
\label{alg:vlm_golden}
\end{figure}

\subsection{Parameter Space}

The fish simulator exposes $d = 16$ parameters spanning three physical subsystems (Table~\ref{tab:params}), which \method{} searches jointly rather than through hand-designed subsystem stages.
The VLM receives parameter descriptions, units, and bounds in the prompt, but no privileged information about the real parameters.

The five fluid coefficients parameterize blunt drag, slender drag, angular drag, Kutta lift, and Magnus lift in a stateless ellipsoid fluid model.
Motor arm length controls the lever arm of the crank-slider tendon mechanism.
Per-joint stiffness and damping govern the mechanical response of each of the five hinge joints in the discretized tail.

\begin{table}[t]
\centering
\caption{Search bounds for the 16 simulator parameters, shared by all methods}
\label{tab:params}
\small
\begin{tabular}{@{}llcc@{}}
\toprule
Parameter group & Count & Bounds & Unit \\
\midrule
Fluid coefficients & 5 & $[0, 10]$ & -- \\
Motor arm length & 1 & $[0.01, 0.06]$ & m \\
Hinge stiffness (per joint) & 5 & $[0.1, 5.0]$ & N$\cdot$m/rad \\
Hinge damping (per joint) & 5 & $[0.0, 2.0]$ & N$\cdot$m$\cdot$s/rad \\
\bottomrule
\end{tabular}
\end{table}

\section{Experimental Setup}
\label{sec:experimental_setup}

\subsection{Hardware Platform}

\begin{figure}[t]
    \centering
    \includegraphics[width=0.85\linewidth]{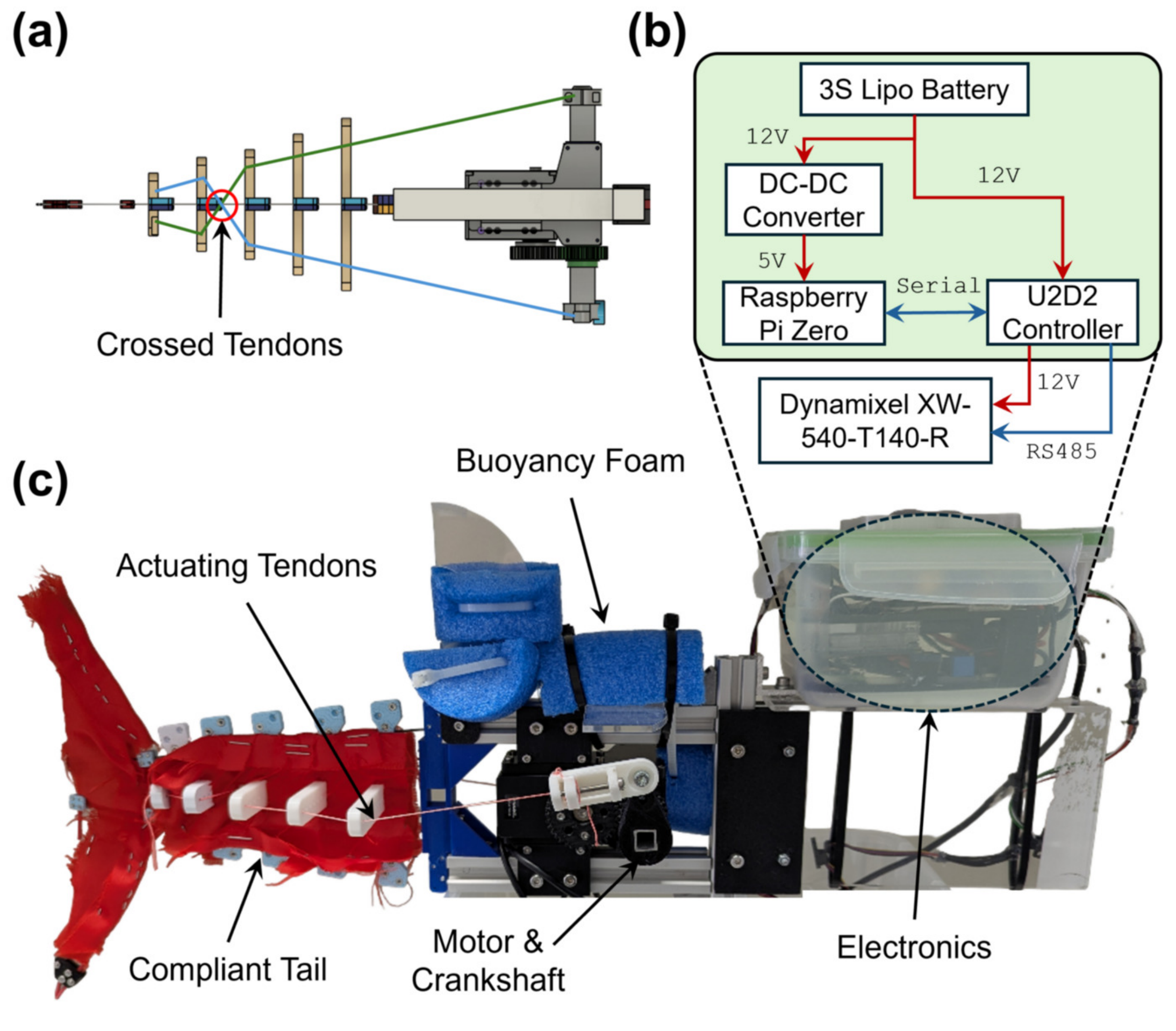}
    \caption{Tendon-driven robotic fish used for calibration and open-loop hardware trials.
    (a)~Computer-aided design (CAD) cross-section showing the antagonistic tendon arrangement that crosses at the tail midpoint to produce an S-bend.
    (b)~Block diagram of the onboard electronics.
    (c)~Assembled robot.}
    \label{fig:hardware}
\end{figure}

The tendon-driven robotic fish is 0.6\,m long and weighs 1.5\,kg including electronics (Fig.~\ref{fig:hardware}).
An onboard computer controls a waterproof servomotor powered by a three-cell lithium-polymer battery (Fig.~\ref{fig:hardware}(b)).

The compliant tail uses super-elastic nitinol rods held by five 3D-printed spine segments, with antagonistic tendons that cross at the midpoint to produce a bio-inspired S-bend (Fig.~\ref{fig:hardware}(a)) driven by a dual-output crank-slider mechanism.

\subsection{Data Collection and Simulation}

Nine centerline body markers are tracked using a discriminative-correlation-filter tracker with channel and spatial reliability (CSRT) from overhead camera footage (2160${\times}$3840, 59.94\,fps) in a 2\,m\,$\times$\,3\,m pool.
For error computation, we rotate the marker trajectories by head orientation and translate them to a common initial origin while retaining forward displacement. Markers M0 and M8 identify the head and tail tip.
The recordings use eight nominal command labels, $\{0.50,0.75,\ldots,2.25\}$\,Hz.
The realized tail-oscillation rates are measured independently from the tracked videos: $\{0.370,0.560,0.749,0.937,1.121,1.309,1.495,1.656\}$\,Hz.
Simulator conditions use these realized rates, with nominal labels retained only as identifiers.

The simulator uses MuJoCo~\cite{todorov2012mujoco} with an implicit integrator at 1\,kHz, discretizing the continuously bending tail as five rigid segments connected by hinge joints.
We sample simulated marker trajectories at explicit physical timestamps and resample them to the real-video time grid, so the comparison uses elapsed time rather than frame index.
Fluid forces use a stateless ellipsoid model parameterized by five coefficients (Table~\ref{tab:params}) that capture the dominant hydrodynamic forces without computational fluid dynamics simulation.
Each evaluation in the main calibration studies averages the marker error over all eight frequencies (Eq.~\ref{eq:objective}).
A single 8-frequency simulation evaluation takes approximately 13\,s on one processor core. We report VLM proposal counts separately because model-query times also depend on network latency and service-side delays.

\subsection{Baselines and Ablations}

We compare eight methods to assess the value of VLM-proposed directions. The two matched controls keep the line search and replace only the direction. BO, SPSA, and CMA-ES provide comparisons with residual-based optimization, while Vid2Sid proposes new parameters for every candidate.
\begin{itemize}
    \item \textit{VLM + line search} is the proposed complete mechanism.
    \item \textit{Multiscale random direction + line search} samples an isotropic unit-Gaussian direction and a normalized radius from $\{0.3,0.6,0.9\}$, then uses the same step sizes $\{1,0.5,0.25\}$.
    \item \textit{Signed-permuted VLM + line search} queries the VLM at that arm's own incumbent, then independently permutes normalized displacement coordinates and applies Rademacher signs before the same line search. This preserves proposal scale and the rule for requesting model proposals while scrambling which physical parameters receive the updates.
    \item \textit{Simultaneous perturbation stochastic approximation (SPSA) + line search} estimates a direction from symmetric normalized probes of radius 0.05 and checks a radius-0.6 proposal with the same backtracking schedule.
    \item \textit{Bayesian optimization} (BO) uses a Gaussian-process surrogate with a 10-point random initial design and an adaptive portfolio of acquisition functions~\cite{snoek2012practical}.
    \item \textit{CMA-ES} uses normalized coordinates, initial standard deviation 0.2, and population sizes 12 and 6~\cite{hansen2016cmaes}.
    \item \textit{Vid2Sid} requests a fresh VLM proposal for each candidate without line search, using the adaptation in Section~\mbox{III-C}.
\end{itemize}

\subsection{Protocol}

The first study uses $B=40$ simulator evaluations for each of fifteen random initializations.
Each seed defines one shared 16-dimensional initialization used by every method, so all comparisons are paired.
The primary metric, area under the best-so-far curve over 40 calls (AUC40), is the mean chronological best-so-far marker error over all 40 calls, including invalid calls that consume budget without improving the incumbent. This captures progress throughout optimization, while final error describes the last best iterate.
The two pre-specified primary contrasts are VLM + line search minus random direction + line search and VLM + line search minus signed-permuted VLM + line search.
We use two-sided exact Wilcoxon signed-rank tests and Holm correction across these two contrasts, and report paired effect sizes with 95\% bias-corrected and accelerated (BCa) intervals.
BO, SPSA, both CMA-ES variants, and Vid2Sid are descriptive secondary comparisons.
The exact prompt template and prompt-builder source accompany this paper as multimedia material. Code, data, and optimization histories will be released on the project page.

After analyzing the fifteen-seed study, we preregistered an independent replication with fifty fresh paired seeds for the complete method and its two matched controls. It uses the same initialization procedure, 40-call budget, evaluator, and statistical tests. The two studies are analyzed separately.
An initialization enters both paired comparisons only when every method returns a valid result, with one shared initial failure retained at the penalty described in Section~V-B.
The paired estimates describe this common subset; completion counts for each method are reported with the replication results.

\section{Results}
\label{sec:results}

\newcommand{\replFullN}{44}
\newcommand{\replRandN}{48}
\newcommand{\replSpN}{45}
\newcommand{\replTripleN}{41}
\newcommand{\replFullExcl}{6}
\newcommand{\replRandExcl}{2}
\newcommand{\replSpExcl}{5}
\newcommand{\replAttritionSentence}{One infeasible initialization was excluded from all methods. A second shared initialization failed identically in all three methods and was retained at the prespecified 1000\,mm penalty. The complete method had four additional incomplete runs (two invalid-response failures and two request failures), and the permuted control had three additional request failures. Exclusions were determined before inspecting outcomes. This leaves 44, 48, and 45 completed runs and 41 initializations completed by all three methods.}
\newcommand{\replFullAuc}{$92.2\pm44.5$}
\newcommand{\replRandAuc}{$159.6\pm79.0$}
\newcommand{\replSpAuc}{$147.4\pm65.6$}
\newcommand{\replFullFinal}{$59.3\pm27.3$}
\newcommand{\replRandFinal}{$84.9\pm72.4$}
\newcommand{\replSpFinal}{$77.1\pm52.8$}
\newcommand{\replRandMeanDiff}{$-63.36$}
\newcommand{\replRandMedDiff}{$-50.10$}
\newcommand{\replRandRB}{$-0.870$}
\newcommand{\replRandBcaMean}{$[-85.29,-44.74]$}
\newcommand{\replRandPraw}{\ensuremath{5.8\times10^{-8}}}
\newcommand{\replRandPholm}{\ensuremath{1.2\times10^{-7}}}
\newcommand{\replSpMeanDiff}{$-53.17$}
\newcommand{\replSpMedDiff}{$-43.31$}
\newcommand{\replSpRB}{$-0.749$}
\newcommand{\replSpBcaMean}{$[-74.22,-32.12]$}
\newcommand{\replSpPraw}{\ensuremath{7.9\times10^{-6}}}
\newcommand{\replSpPholm}{\ensuremath{7.9\times10^{-6}}}
\newcommand{\replNpairs}{42}
\newcommand{\replClassificationSentence}{Both preregistered contrasts pass Holm correction in the replication. The earlier fifteen-seed study remains non-confirmatory.}

\newcommand{\replDiagN}{41}
\newcommand{\replFinalRandomMeanDiff}{-25.2}
\newcommand{\replFinalRandomCi}{[-53.1,-4.8]}
\newcommand{\replFinalRandomWins}{31}
\newcommand{\replFinalPermutedMeanDiff}{-16.0}
\newcommand{\replFinalPermutedCi}{[-34.6,-2.4]}
\newcommand{\replFinalPermutedWins}{29}
\newcommand{\replMarkerHead}{41.4}
\newcommand{\replMarkerMid}{62.1}
\newcommand{\replMarkerTail}{77.2}
\newcommand{\replMarkerTip}{90.4}

\newcommand{\directN}{15}
\newcommand{\directAuc}{74.1\pm16.4}
\newcommand{\directFinal}{52.1\pm14.3}
\newcommand{\directCallsMean}{18.0}
\newcommand{\directCallsMedian}{10}

\newcommand{\interpFullMeanSd}{53.1\pm3.8}
\newcommand{\interpRandomMeanSd}{102.2\pm64.9}
\newcommand{\interpBoMeanSd}{65.4\pm28.5}
\newcommand{\interpReferenceMeanSd}{61.6\pm22.1}
\newcommand{\interpRandomMeanDiff}{-49.1}
\newcommand{\interpRandomMedianDiff}{-16.6}
\newcommand{\interpRandomMeanCi}{[-104.0,-16.5]}
\newcommand{\interpRandomWins}{7}
\newcommand{\interpBoMeanDiff}{-12.3}
\newcommand{\interpBoMedianDiff}{-5.8}
\newcommand{\interpBoMeanCi}{[-41.8,0.6]}
\newcommand{\interpBoWins}{6}
\newcommand{\interpReferenceMeanDiff}{-8.5}
\newcommand{\interpReferenceMedianDiff}{0.6}
\newcommand{\interpReferenceMeanCi}{[-32.6,0.6]}
\newcommand{\interpReferenceWins}{3}
\newcommand{\interpNpairs}{8}

\subsection{Fifteen-Seed Matched-Control Study}

Across fifteen shared initializations, \method{} reaches a mean AUC40 of $93.5\pm40.8$\,mm and mean final error of $59.9\pm34.2$\,mm (Table~\ref{tab:confirmatory}). Two runs end above 100\,mm, while the median final error is 46.7\,mm. Vid2Sid attains lower mean AUC40 ($\directAuc$\,mm) and final error ($\directFinal$\,mm), using 39 proposals per run compared with 16.1 for \method{}.

\begin{table*}[t]
\centering
\caption{\textbf{Fifteen-seed calibration comparison.} All methods use 40 simulator evaluations per shared initialization. AUC40 is the mean best-so-far error over evaluations. Errors are mean $\pm$ standard deviation (SD) over runs. Threshold calls include every run, assigning 41 when 55\,mm is not reached. Failure totals count failed evaluations / failed single-frequency simulations. Bold marks the lowest observed mean error or threshold count, not statistical significance. Secondary comparisons are descriptive. Vid2Sid uses 2.4 times as many proposals as \method{}}
\label{tab:confirmatory}
\small
\setlength{\tabcolsep}{5pt}
\begin{tabular}{lcccrr}
\toprule
Method & AUC40 $\downarrow$ & Final error $\downarrow$ & Calls to 55\,mm $\downarrow$ & VLM proposals & Failure totals \\
 & (mm) & (mm) & capped mean (median) & mean per run & calls / frequencies \\
\midrule
Vid2Sid & $\bestMeanSd{\directAuc}$ & $\bestMeanSd{\directFinal}$ & \textbf{\directCallsMean} (\directCallsMedian) & 39.0 & 0/0 \\
\rowcolor{black!12}\method{} (Ours) & $93.5\pm40.8$ & $59.9\pm34.2$ & 20.9 (16) & 16.1 & 0/0 \\
BO & $114.2\pm56.6$ & $60.8\pm38.7$ & 21.7 (20) & 0 & 11/31 \\
SPSA + line search & $123.5\pm52.4$ & $66.8\pm44.7$ & 22.2 (16) & 0 & 1/1 \\
CMA-ES (pop. 6) & $124.7\pm41.6$ & $54.7\pm8.2$ & 32.1 (34) & 0 & 4/5 \\
Permuted VLM + line search & $128.4\pm54.6$ & $60.5\pm21.3$ & 30.8 (37) & 15.9 & 19/53 \\
Random + line search & $140.2\pm74.8$ & $92.7\pm77.9$ & 26.9 (33) & 0 & 6/9 \\
CMA-ES (pop. 12) & $140.6\pm68.8$ & $78.4\pm52.6$ & 29.6 (41) & 0 & 5/8 \\
\bottomrule
\end{tabular}
\end{table*}

The paired mean AUC40 difference (\method{} minus control) is $-46.66$\,mm against random directions (unadjusted 95\% BCa interval $[-97.25,-4.74]$\,mm) and $-34.88$\,mm against signed-permuted directions ($[-65.95,-1.08]$\,mm). Exact two-sided Wilcoxon tests give Holm-adjusted $p=0.1205$ and $0.0825$, respectively. Neither contrast passes the preregistered familywise threshold, so the fifteen-seed study is non-confirmatory.

\begin{figure}[t]
\centering
\includegraphics[width=\columnwidth]{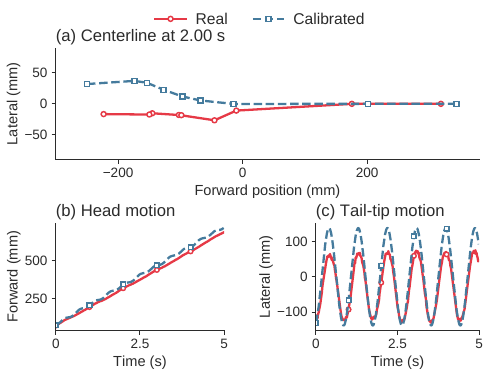}
\caption{Recorded and calibrated motion for the initialization with median final error in Table~\ref{tab:confirmatory}, at the nominal 1.50\,Hz fitting condition (realized 1.121\,Hz). (a) Nine centerline markers at 2.00\,s. (b) Forward head motion. (c) Lateral tail-tip motion. All panels use the same aligned evaluation window. Mean marker error at this condition is 34.8\,mm.}
\label{fig:calibrated_motion}
\end{figure}

Fig.~\ref{fig:calibrated_motion} shows recorded and calibrated motion for the initialization with median final error. The centerline snapshot and time traces separate forward progression from lateral tail motion, both of which contribute to the marker objective.

\subsection{Independent Fifty-Seed Replication}

\begin{figure}[t]
\centering
\includegraphics[width=\columnwidth]{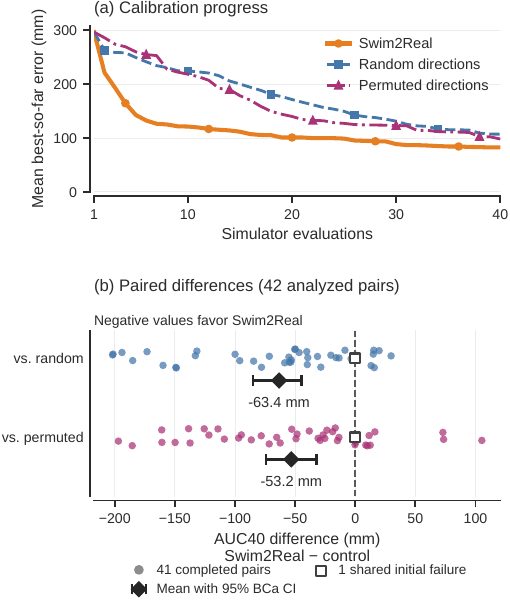}
\caption{VLM directions reduce calibration error relative to matched controls in the independent replication. (a) Mean best-so-far error over \replNpairs{} analyzed initializations, including the shared 1000\,mm failure. (b) Paired AUC40 differences for the \replNpairs{} analyzed initializations, with paired means and unadjusted 95\% BCa intervals. The hollow square marks the shared failure's zero difference. Both Holm-adjusted tests pass. Apart from this shared failure, the analysis conditions on joint completion.}
\label{fig:replication}
\end{figure}

The replication compares the complete method and both matched controls on fifty fresh paired seeds (Table~\ref{tab:replication}, Fig.~\ref{fig:replication}). The paired analysis includes the \replTripleN{} initializations completed by all methods and one shared initialization that failed identically in all three methods. This common failure receives a 1000\,mm penalty in each method and contributes a zero paired difference.

\begin{table}[t]
\centering
\caption{\textbf{Fifty-seed replication.} Top: mean $\pm$ SD over $n$ completed runs per method. Bottom: paired AUC40 differences (\method{} minus control) on \replTripleN{} jointly completed initializations plus one common failure. Paired estimates use this common subset rather than subtracting the means above. Lower values favor \method{}. Intervals are unadjusted 95\% BCa, while $p$-values are Holm-adjusted exact Wilcoxon tests}
\label{tab:replication}
\footnotesize
\setlength{\tabcolsep}{2pt}
\begin{tabular}{lccc}
\toprule
\multicolumn{4}{l}{\textbf{Per-method summaries}} \\
Method & $n$ & AUC40 & Final error \\
 & & (mm) & (mm) \\
\midrule
\rowcolor{black!12}\method{} (Ours) & \replFullN & \replFullAuc & \replFullFinal \\
Random directions & \replRandN & \replRandAuc & \replRandFinal \\
Permuted directions & \replSpN & \replSpAuc & \replSpFinal \\
\midrule
\multicolumn{4}{l}{\textbf{Paired AUC40 differences} (\replNpairs{} initializations)} \\
Control & Mean $\Delta$ (mm) & 95\% interval (mm) & Holm $p$ \\
\midrule
Random & \replRandMeanDiff & \replRandBcaMean & \replRandPholm \\
Permuted & \replSpMeanDiff & \replSpBcaMean & \replSpPholm \\
\bottomrule
\end{tabular}
\end{table}

Among the 41 jointly completed initializations, \method{} has lower AUC40 in 34 pairs against random and 32 against permuted directions, with both contrasts passing Holm correction (Table~\ref{tab:replication}).

On the same \replDiagN{} jointly completed seeds, a post hoc descriptive comparison gives paired mean final-error differences of $\replFinalRandomMeanDiff$\,mm against random directions (95\% BCa $\replFinalRandomCi$\,mm, \replFinalRandomWins/\replDiagN{} wins) and $\replFinalPermutedMeanDiff$\,mm against permuted directions ($\replFinalPermutedCi$\,mm, \replFinalPermutedWins/\replDiagN{} wins).

\Needspace{3\baselineskip}
\subsection{Second-Model Experiment}
\label{sec:second_model}

With GLM-5.3-Flash~\cite{zai2026glm53flash}, we evaluated 24 paired initializations using 16 sampled frames per clip and the same simulator budget and line search. We ran the complete and permuted arms with GLM and reused the random-direction runs from the replication, since that control queries no model.
The complete method attained lower AUC40 than both controls (Table~\ref{tab:second_model}). Its median AUC40 was 71.1\,mm, compared with 161.1\,mm for random and 113.2\,mm for permuted directions. This second-model result supports the same direction effect observed with Gemini.

\begin{table}[t]
\centering
\caption{\textbf{Second-model paired comparison.} AUC40 differences (\method{} minus control) with GLM-5.3-Flash on 24 paired initializations. Negative favors \method{}. Intervals are unadjusted 95\% BCa, with exact Wilcoxon $p$-values Holm-adjusted over the two comparisons. Random-direction runs are reused from the replication}
\label{tab:second_model}
\footnotesize
\setlength{\tabcolsep}{2pt}
\begin{tabular}{lccc}
\toprule
Control & Mean $\Delta$ (mm) & 95\% interval (mm) & Holm $p$ \\
\midrule
Random & $-88.3$ & $[-121.2,-58.7]$ & $1.7\times10^{-5}$ \\
Permuted & $-58.6$ & $[-87.1,-35.0]$ & $9.1\times10^{-5}$ \\
\bottomrule
\end{tabular}
\end{table}

\subsection{Eight-Seed Frequency Interpolation}
\label{sec:interpolation}

For eight fresh paired initializations, the complete method, random directions, and BO fit nominal $\{0.50,1.00,1.50,2.00,2.25\}$\,Hz for 40 calls. Objectives, VLM inputs, diagnostics, history, and selection use only these conditions. We freeze the training-best parameters before one assessment at $\{0.75,1.25,1.75\}$\,Hz. The all-frequency context arm fits all eight conditions at greater per-call simulation cost. The recordings were already used during development.

The complete method wins \interpRandomWins/8 pairs against random directions and \interpBoWins/8 against BO (Table~\ref{tab:interpolation}). The BO mean interval includes zero. These seed-resampled intervals quantify optimizer variation on the same recordings.

The all-frequency reference shows how fitting all eight conditions affects error at the three assessment frequencies under the same 40-call budget. Because its objective also rewards accuracy at the other five conditions, it provides context rather than an upper bound on interpolation accuracy. The paired mean difference (\method{} minus reference) is $\interpReferenceMeanDiff$\,mm with a descriptive 95\% interval of $\interpReferenceMeanCi$\,mm.
\begin{table}[t]
\centering
\caption{\textbf{Eight-seed frequency interpolation.} Recordings were development-exposed. Error is assessed at three frequencies excluded from fitting, except in the all-frequency reference. Top: mean $\pm$ SD. Bottom: paired error differences (\method{} minus comparator), with descriptive 95\% seed-resampled BCa intervals, without significance tests. Negative favors \method{}. Bold marks the lowest observed interpolation mean}
\label{tab:interpolation}
\footnotesize
\setlength{\tabcolsep}{3pt}
\begin{tabular}{lcc}
\toprule
Method & Frequencies fit & Assessment error \\
 & & (mm) \\
\midrule
\rowcolor{black!12}\method{} (Ours) & 5 & $\bestMeanSd{\interpFullMeanSd}$ \\
Random + line search & 5 & $\interpRandomMeanSd$ \\
BO & 5 & $\interpBoMeanSd$ \\
\midrule
\multicolumn{3}{l}{\textbf{All-frequency reference} (fits assessment frequencies)} \\
All-frequency context & 8 & $\interpReferenceMeanSd$ \\
\midrule
\multicolumn{3}{l}{\textbf{Paired error differences} (eight seeds)} \\
Comparator & Mean $\Delta$ (mm) & 95\% interval (mm) \\
\midrule
Random & $\interpRandomMeanDiff$ & $\interpRandomMeanCi$ \\
BO & $\interpBoMeanDiff$ & $\interpBoMeanCi$ \\
\bottomrule
\end{tabular}
\end{table}

\subsection{Open-Loop Hardware Deployment}

\begin{figure}[t]
\centering
\includegraphics[width=\columnwidth]{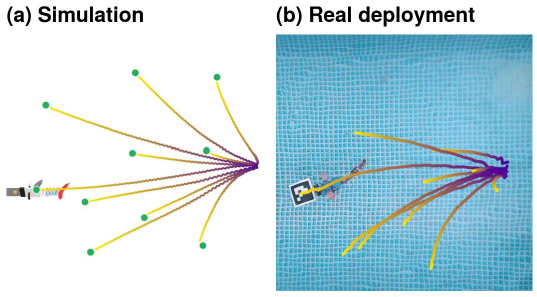}
\caption{Open-loop deployment of simulator-generated motor commands.
(a) Simulated policy rollouts with goal positions marked. (b) Recorded hardware motion during open-loop replay of the same command sequences.
The trials assess displacement during open-loop replay, without feedback or a matched hardware control. The policies were trained with an earlier simulator.}
\label{fig:deployment_scope}
\end{figure}

We replayed motor commands from soft actor-critic policies~\cite{haarnoja2018soft} in ten open-loop hardware trials (Fig.~\ref{fig:deployment_scope}). The policies were trained with an earlier simulator, separately from the calibration studies reported here. All ten sequences executed with positive signed initial-heading displacement, averaging $+0.382\pm0.220$\,m, and mean net displacement was $0.566\pm0.344$\,m. These trials establish the feasibility of executing simulator-generated commands on the robot.

\Needspace{4\baselineskip}
\section{Discussion}
\label{sec:discussion}

\subsection{Calibration Error and Model-Query Cost}

\onelinelonger{}The replication and the second-model experiment identify useful information in the VLM's proposed directions under a shared line search and simulator budget. The 40-call budget includes evaluations of both new proposals and smaller steps along an existing direction. Backtracking allocates part of that budget to refining proposals, while Vid2Sid uses every call after initialization to evaluate a fresh proposal. Backtracking averages 16.1 VLM proposals per run. Vid2Sid uses 39 proposals and attains lower error at the same simulator budget (Table~\ref{tab:confirmatory}), so the choice depends on the relative cost of model queries and residual calibration error. Reusing video-based proposals reduces query demand when model calls are rate limited, billed per token, or slow.

In a post hoc profile of the 242 recorded proposal rounds in the fifteen-seed study, the complete method accepted 97 updates: 51 at the full step, 29 at the half step, and 17 at the quarter step. Backtracking therefore supplied 46 of the 97 accepted updates (47\%), each using a direction whose full step had not improved the incumbent. Accepted updates reduced marker error by a median of 14.5\,mm.

We also replayed each recorded proposal for one round under nine schedules with $\beta\in\{0.25,0.5,0.75\}$ and $K\in\{2,3,4\}$. With step halving, a fourth step would have accepted 16 further proposals for 145 additional evaluations. Its median improvement was 0.29\,mm, compared with 9.1\,mm at the third step. This replay compares step choices at fixed incumbents. Comparing complete runs within a 40-call budget would also require propagating accepted updates into later VLM proposals.

\subsection{Calibration Error Structure}

\onelinelonger{}The mean final marker error is about 60\,mm in both studies, or 10\% of the 600\,mm body length. We measure error along the body centerline, whereas prior differentiable calibration~\cite{zhang2022sim2real} reports actuator-length error. A direct comparison would require the same measured quantity and normalization.
In a post hoc decomposition of the final iterates on the \replDiagN{} jointly completed replication seeds, \method{} has mean errors of \replMarkerHead\,mm at the head (M0--M2), \replMarkerMid\,mm at mid-body (M3--M5), and \replMarkerTail\,mm at the tail (M6--M8), reaching \replMarkerTip\,mm at the tip. Both controls have higher mean error in each region, with residuals increasing toward the tail.

\Needspace{3\baselineskip}
\subsection{Scope of VLM-Guided Calibration}

\onelinelonger{}The replication tests whether retaining the VLM's proposed directions improves calibration under a shared line search. Random directions measure the progress attainable with the same line search and no model queries. Signed permutation preserves the proposed displacement norm in normalized coordinates before projection onto the bounds. Expressing each coordinate relative to its search range allows updates to motor arm length, fluid coefficients, stiffness, and damping to be reassigned despite their different units. This control tests whether the VLM's assignment of relative changes to named parameters improves on retaining the proposal's scale. Both arms receive the same objective-based acceptance \mbox{feedback}.

Each method updates its own parameters, so the paired results compare the cumulative effect of direction proposals along different optimization trajectories. The post hoc final-error comparisons follow the same pattern (Section~\mbox{V-B}). The broader optimizer comparisons in Table~\ref{tab:confirmatory} remain descriptive.

\subsection{Hardware Feasibility}

The hardware trials establish open-loop command execution using policies from an earlier simulator. Testing the calibration reported here on hardware requires feedback, measured target attainment, and a matched control.

\subsection{Limitations}

\onelinelonger{}The recovered vector matches behavior over eight observed actuation frequencies rather than uniquely identifying material or fluid parameters. The interpolation split withholds three conditions from fitting but uses recordings seen during development, and seeds vary optimizer initialization on the same fish, so new recordings and swimming trajectories are future work.
The five-link tail approximation limits continuous bending. The prompt combines video with numerical residuals, so the separate contribution of video and the correctness of the model's physical explanations are open questions.
The replication used Gemini 2.5 Pro, with the GLM experiment as second-model evidence.
The descriptive Vid2Sid comparison measures error and proposal count. Evaluating equal query or wall-clock budgets is future work.

\section{Conclusion}
\label{sec:conclusions}

\method{} combines VLM-proposed parameter directions with objective-checked step sizes to calibrate a 16-parameter fish simulator from swimming video. Under a shared line search and simulator budget, these directions reduce calibration error relative to random and signed-permuted directions in a fifty-seed replication and with a second VLM. An eight-seed descriptive study also finds lower mean error at frequencies withheld from fitting. The VLM proposes physical hypotheses, and the simulator objective determines which step sizes improve calibration. Next steps are hardware evaluation with feedback and a matched control, and comparison of reused and fresh proposals at equal query budgets.

\section*{Acknowledgment}
The authors thank the Faculty of Mathematics, Informatics and Mechanics of the University of Warsaw for access to its Entropy computing cluster.

\balance

\bibliographystyle{IEEEtran}
\bibliography{references}

\end{document}